\documentclass[a4paper]{article}

\usepackage{INTERSPEECH2021}
\usepackage{multirow}
\usepackage{multicol}
\usepackage{url}
\usepackage[main=english,vietnamese]{babel}
\usepackage[utf8]{inputenc}
\title{Vietnamese Capitalization and Punctuation Recovery Models}
\name{Hoang Thi Thu Uyen$^1$, Nguyen Anh Tu$^1$, Ta Duc Huy$^2$ }
\address{$^1$Posts and Telecommunications Institute of Technology, Hanoi, Vietnam \\
$^2$Independent Researcher, Ho Chi Minh City, Vietnam}
\email{\{thuuyenptit,anhtunguyen446,tdh512194\}@gmail.com}

\begin{document}

\maketitle
\begin{abstract}

Despite the rise of recent performant methods in Automatic Speech Recognition (ASR), such methods do not ensure proper casing and punctuation for their outputs.
This problem has a significant impact on the comprehension of both Natural Language Processing (NLP) algorithms and human to process.
Capitalization and punctuation restoration is imperative in pre-processing pipelines for raw textual inputs.
For low resource languages like Vietnamese, public datasets for this task are scarce.
In this paper, we contribute a public dataset for capitalization and punctuation recovery for Vietnamese; and propose a joint model for both tasks named JointCapPunc.
Experimental results on the Vietnamese dataset show the effectiveness of our joint model compare to single model and previous joint learning model.
We publicly release our dataset and the implementation of our model at {\url{https://github.com/anhtunguyen98/JointCapPunc}}

\end{abstract}

\section{Introduction}
Generic ASR systems only transcribe audio input to raw textual output of single-lowercase word sequences.
For a given output sequence of an ASR system: \textit{'hi uyen how are you'} , the properly capitalized and punctuation recovered output would be: \textit{'Hi Uyen, how are you?'}.
Restoring punctuation and capitalization enormously improves the readability of ASR output.
This step also improves the results for common subsequent tasks in NLP like Named entity recognition, Intent classification, etc.
\cite{cho2012segmentation, nguyen20d_interspeech, tkachenko2012named} 
Data resources for Vietnamese capitalization and punctuation recovery is limited. There are only two public datasets \cite{pham2020vietnamese} for punctuation prediction and no available dataset for capitalization restoration.
These two datasets are aggregated from news articles and novels, which do not contain more context on spoken language
, thus do not generalize well enough for applications in ASR systems.

In this work, we aim to apply capitalization and punctuation recovering to ASR systems in the medical domain.
Our method can enhance the readability of ASR-transcribed records of the conversations between COVID-19 patients and doctors.

We present a large-scale public dataset for Vietnamese capitalization and punctuation recovering focus on the medical domain.
The dataset contains context on spoken language that consist thousands of question and answer pairs that crawled from healthcare websites.
We consider common punctuation marks and capitalization types presented in section \ref{sec:dataset}.

Traditional methods often solve these tasks independently as two sequence labeling models.
These methods exploit probabilistic graphical models like conditional random fields (CRFs) \cite{10.5555/645530.655813} \cite{wang2006capitalizing} \cite{pham2019punctuation} and combine with common neural networks such as CNNs and LSTM \cite{elasko2018PunctuationPM} \cite{pham2020vietnamese} \cite{tran2021neural}. 
For Vietnamese, recently pre-trained language models \cite{devlin2019bert} have made breakthroughs in various Vietnamese NLP tasks and applications including analyzing questions \cite{anh2021analyzing}, punctuation prediction \cite{tran2021punctuation}, sentiment analysis \cite{nguyen2020fine}, sequence tagging \cite{tran2020improving}.
In this paper, we proposed a new joint learning model
that based on pre-trained transformers language models and extend with a capitalization prediction layer to produce the soft capitalization features to combine with contextual features for the punctuation prediction layer.
Our contributions are summarized as follows:

\begin{itemize}
\item We introduce a Vietnamese public capitalization and punctuation recovery dataset—named \textbf{ViCapPunc}.
\item We proposed a capitalization and punctuation recovery model named \textbf{JointCapPunc}. Experimental results on our dataset
show the benefits of joint learning over learning the model separately on each task.
\end{itemize}
\section{Related Work}
In addition to IQ2\footnote{\url{http://www.intelligencesquaredus.org}}, IWSLT 2011 \cite{federico2012iwslt} are common datasets which used for capitalization and punctuation prediction in English.
All of these text are the transcripts of the conversations that contain spoken language.
To our knowledge, there are two public datasets et al \cite{pham2020vietnamese} for punctuation prediction task which on news domain and novels domain in Vietnamese.
However two Vietnamese dataset do not focus on spoken language domain that make improvement for ASR system.
Besides, these dataset do not consist annotation for capitalization prediction task.

Early approaches to solve the two tasks are using single sequence labeling model including n-gram language model \cite{gravano2009restoring}, conditional random fields model \cite{pham2019punctuation} \cite{wang2012dynamic}, deep neural models \cite{pham2020vietnamese}, \cite{tran2021punctuation}. Recently, there are some studies proposed to solve the two tasks simultaneously following two strategies.
The first strategy is combining annotations of capitalization and punctuation task and formulate the task as sequence labeling \cite{nguyen2021toward} or sequence to sequence models \cite{nguyen2019fast}.
The second strategy is to jointly learn a model for both tasks, which performs better than its single counterpart .
\cite{pahuja2017joint} employs bidirectional recurrent neural networks and two independent prediction layers
, which does not leverage capitalization features for punctuation prediction tasks.
Monica et al \cite{dixit2020robust} proposed a method that conditions truecasing prediction on punctuation output. They concatenate the softmax probabilities of punctuation output with BERT encoder’s outputs and feed to a linear to predict truecasing. Our method is similar to this work but we use the output probability of capitalization head to produce soft capitalization features.
We then combine these features with the contextual word features to predict punctuation marks.
Besides, to improve performance of punctuation task, we extend model with conditional random field layer. This model is presented in detail in section \ref{sec:method}.

\section{ViCapPunc Dataset}
\label{sec:dataset}
To investigate the two tasks
for Vietnamese.
We build a large scale dataset namely ViCapPunc that contains spoken language from the medical domain.
We first crawl 50,000 question and answer pairs from the reputable Vietnamese healthcare sites online including Vinmec\footnote{\url{https://www.vinmec.com}}, Alobacsi\footnote{\url{https://alobacsi.com}} and ISOFHCARE\footnote{\url{https://isofhcare.com}}.
There are 425,350 sentences in the ViCapPunc dataset.
We define two types of label for the capitalization prediction task including: word with the first character capitalized (CAP), word with all capitalized characters (ALL-CAP) and three common punctuation marks for the punctuation prediction task including the comma(,), the period(.) and the question mark (?).

We split 50,000 question and answer pairs into training, validation and test sets randomly with the ratio 50\%, 20\%, 30\% respectively. 
For pre-processing steps, we converted all data into lower case, removed the special characters and punctuation marks, keeping commas, periods and question marks.
Finally, we label the dataset following the format in table \ref{tab:format} that consists of three columns.
Column 1 is the lower case text of input sentences.
Column 2 is the capitalization label, where 0 denotes non-capitalized word, 1 is CAP and 2 is ALL-CAP.
Columns 3 is the punctuation label where O indicates that a word is not followed by any punctuation.
Following the strategy in Thuy et. al. \cite{pham2020vietnamese} where the maximum length of a question and answer is 150 words, we do not segment data into paragraphs. Instead, 
we decided to split the data into segments of length equal to 150 and made sure that the last word of a segment input is the last word of a sentence.
Table \ref{tab:stat-dataset} shows the statistics of our dataset.
\begin{table}[]
\centering
\caption{The format of the dataset.}
\vspace{-0.3cm}
\begin{tabular}{lllll}
\textbf{Text}  &  & \textbf{Cap label} &  & \textbf{Punctuation label}     \\
chào  &  & 1 &  & O     \\
uyên  &  & 1 &  & COMMA \\
bạn   &  & 0 &  & O     \\
có    &  & 0 &  & O     \\
khoe  &  & 0 &  & O     \\
không &  & 0 &  & QMARK
\end{tabular}
\label{tab:format}
\end{table}

\begin{table}[]
\centering
\caption{Statistics of ViCapPunc dataset.}
\vspace{-0.3cm}
\begin{tabular}{l|l|l|l}
\hline
\textbf{Punctuation}    & \multicolumn{1}{c|}{\textbf{Train}} & \multicolumn{1}{c|}{\textbf{Validation}} & \textbf{Test} \\ \hline
COMMA        &  228,495   &      92,026        &    137,865           \\
PERIOD       &  184,383   &      74,097        &    111,717             \\
QMARK        &  27,448    &      10,987        &    16718           \\ \hline
\textbf{Capitalization} &                      &                &               \\ \hline
CAP          &  318,654    &      127,222         &       191,767        \\
ALL-CAP      &  37,542     &      14,582         &        23,173       \\ \hline
\textbf{Sentences}      &       211,831      &        85,084       &        128,435       \\ \hline
\end{tabular}
\label{tab:stat-dataset}
\end{table}

\section{JointCapPunc}
\label{sec:method}
\begin{figure*}[t]
    \centering
    \includegraphics[width=0.8\textwidth]{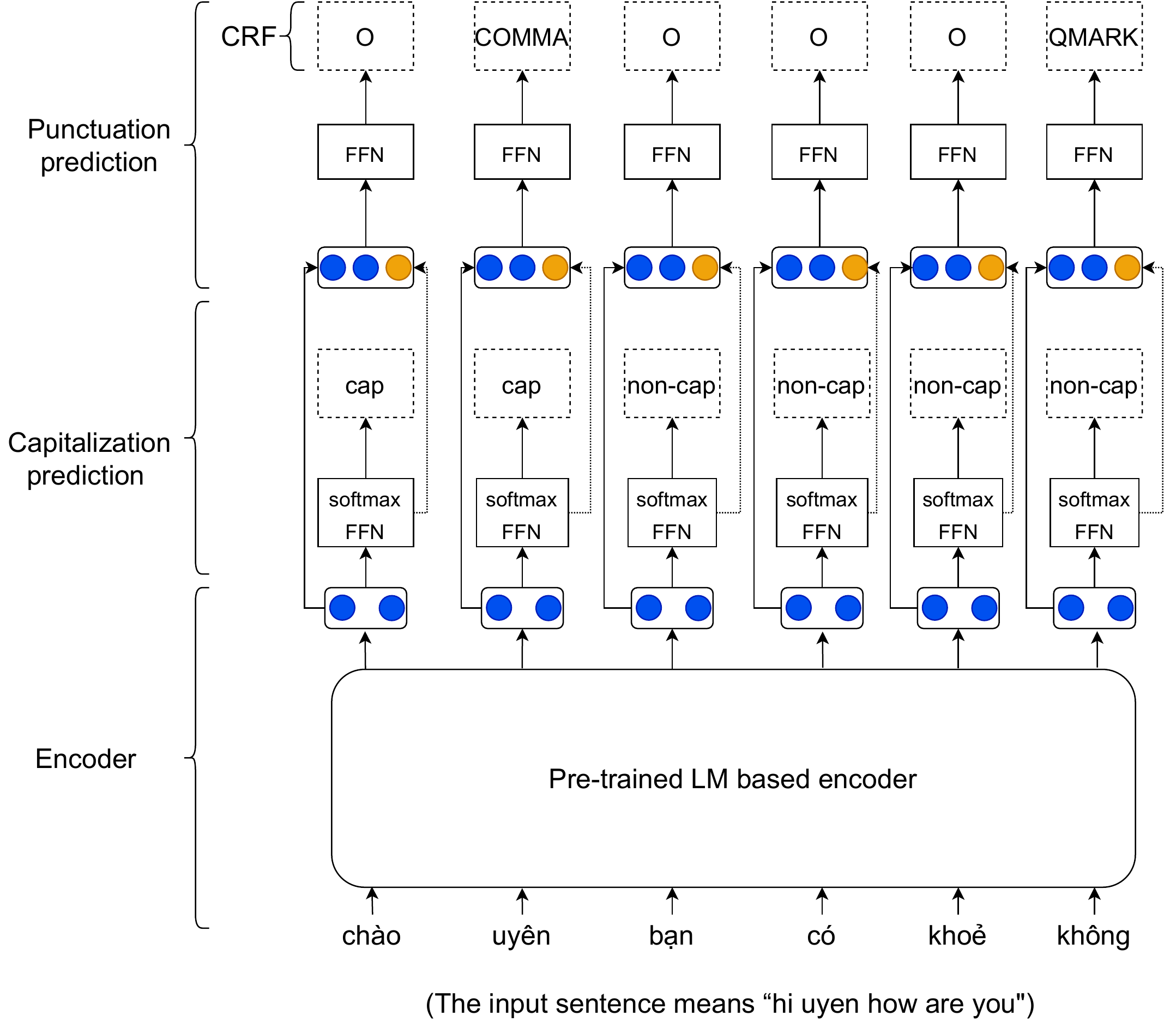}
    \caption{Illustration of our JointCapPunc model.}
    \label{fig: model}
\end{figure*}

Figure 1 illustrates the architecture of our joint model—named \textbf{JointCapPunc}, which consists of three layers including: an encoding layer (i.e. encoder), and two decoding layers of capitalization prediction and punctuation prediction\\
\textbf{Encoding layer:} Given an input sentences consist of n words $w_1,w_2 ... w_n$.
The encoding layer employs a pre-trained language model based on Transformers encoder \cite{vaswani2017attention} like RoBERTa \cite{liu2019roberta}, Electra \cite{clark2020electra}.
Because RoBERTa, Electra segments the input sentence to sub-words which enables the encoding of rare words with appropriate sub-words.
We re-combined sub-word representation after encoding layer into word representation using sum operation. The $i^{th}$ word $w_i$ is represented by vector $e_i$ as follows:

\begin{equation}
    e_i = PretrainedLM(w_{1:n},i)
\end{equation}
\textbf{Capitalization prediction layer:}
Following the common approached when fine-tuning a pre-trained language model for token classification task \cite{devlin2019bert}, the punctuation prediction layer is a linear prediction layer which is added on the top of encoder layer.
In particular, we used word vector representation $e_i$ that is fed into feed-forward network layer ($FFN_{cap}$) followed by softmax predictor for capitalization prediction.

\begin{equation}
    p_i = softmax(FFN_{cap}(e_i)),
\end{equation}
where the output size of $FFN_{cap}$ is 3, including non-capitalized words, words with first character capitalized, words with all capitalized characters.
The cross entropy loss function $\mathcal{L}_{cap}$ is used to optimize capitalization prediction.\\
\\\textbf{Punctuation prediction layer:} The punctuation prediction task is formulated as sequence labeling problem. First, we create a sequence of vector $x_{1:n}$ in which $x_i$ is the concatenation of the word representation $e_i$ and the soft capitalization embedding $c_i$.
\begin{equation}
    x_i = e_i \oplus c_i,
\end{equation}
where the soft capitalization embedding is computed by multiplying the weight matrix $W$ with the corresponding output probability vector $p_i$
\begin{equation}
    c_i = W p_i.
\end{equation}
Then each vector $x_i$ is fed into the $FFN_{punc}$ layer:
\begin{equation}
    s_i = FFN_{punc}(x_i),
\end{equation}
where the output size of $FFN_{punc}$ layer is the number of punctuation label.
The punctuation prediction layer feeds output vector $s_i$ into a conditional random field layer \cite{10.5555/645530.655813}.
Negative log likelihood loss $\mathcal{L}_{punc}$ is used for punctuation prediction task optimization.\\
\\\textbf{Joint training:} The final training objective loss $\mathcal{L}$ of our proposed JointCapPunc model is the weighted sum of the capitalization prediction loss $\mathcal{L}_{cap}$ and punctuation prediction loss $\mathcal{L}_{punc}$
\begin{equation}
    \mathcal{L} = \lambda \mathcal{L}_{cap} + (1-\lambda) \mathcal{L}_{punc}
\end{equation}

\section{Experiments}
\subsection{Evaluation metrics}
The performance of our model is evaluated by using precision, recall and f1-score for both task.
\begin{equation}
    precision = {TP\over{TP+FP}},
\end{equation}
\begin{equation}
    recall = {TP\over{TP+FN}},
\end{equation}
\begin{equation}
F_1 = {2*precision*recall\over{precision+recall}},
\end{equation}
where $TP$ (true positive) is the number of punctuation mark / capitalization mark correct predictions. $FP$ (false positive) is the number of punctuation mark / capitalization mark that are falsely predicted. $FN$ (false negative) is the number of punctuation marks / capitalization marks that are not identified.
\subsection{Experimental Setup}

We conduct experiments on our ViCapPunc dataset to study the effectiveness of joint training model. Here, we employ XLM-R \cite{conneau2020unsupervised} is pre-trained on a 2.5TB multilingual dataset that contains 137 GB Vietnamese text and vELECTRA \cite{tran2020improving}, a monolingual variant of ELECTRA for Vietnamese, is pre-trained on 10GB of Vietnamese text.

Our models were implemented in PyTorch \cite{paszke2019pytorch} using Huggingface's Transformers \cite{wolf-etal-2020-transformers}. Both single-task learning models and jointly learning model were trained using the AdamW optimizer \cite{loshchilov2018decoupled} and set the batch size to 32. We set the epsilon and weight decay to default value in PyTorch, i.e. 1e-8, the soft capitalization embedding dim is 256 . The learning rate and mixture weight $\lambda$ were tuned in $\{$1e-5, 2e-5, 3e-5, 4e-5, 5e-5$\}$ and $\{$ 0.05, 0.1, 0.15 ... 0.95$\}$ respectively. The best learning rate and mixture weight value was 5e-5 and 0.15 respectively. We train for 10 epochs and calculate the average score of the $F_1$-score for capitalization prediction and punctuation prediction after each training epoch on validation set and selected version that obtained the highest average score on the validation set to apply to the test set.

\subsection{Experimental Results}
Table \ref{tab:result} shows results obtain for our JointCapPunc model compares with baseline model of single task training and joint model Monica et al \cite{dixit2020robust}.
For single task training model: Following common strategy when fine-tuning a pre-trained language model for token classification task \cite{devlin2019bert}. We append a linear prediction layer on the top of pre-trained language models as briefly described in Section \ref{sec:method} which is applied for the capitalization prediction task. For punctuation prediction task, we use a punctuation restore model for Vietnamese \cite{tran2021efficient} that append a CRF prediction layer on top of pre-trained language models. The single training strategy based on the XLM-R and vELECTRA
In each setting, JointCapPunc model performs better than single training approach and joint model Monica et al \cite{dixit2020robust} on both task.
Here, the highest improvements are accounted for the punctuation $F_1$ (ie., 78.65\% $\to$ 79.33\% $\to$ 80.10\%) and the capitalization $F_1$ (ie., 83.55\% $\to$ 84.16\% $\to$ 86.75\%).
Thus showing that our joint training model and soft capitalization features helps to better exploit the relationship between the two tasks.
Besides, vELECTRA outperforms XLM-R (punctuation mirco-$F_1$: 80.10\% vs 78.74\%, capitalization micro-$F_1$: 86.75\% vs 85.39\%).

\begin{table*}[]
\centering
\caption{Experimental result (in \%) on testset }
\vspace{-0.3cm}
\begin{tabular}{l|l|ccc|ccc}
\hline
\multicolumn{1}{c|}{\multirow{2}{*}{Encoder}} & \multicolumn{1}{c|}{\multirow{2}{*}{Model}} & \multicolumn{3}{c|}{Capitalization}                                                    & \multicolumn{3}{c}{Punctuation}                                                       \\ \cline{3-8} 
\multicolumn{1}{c|}{}                         & \multicolumn{1}{c|}{}                       & \multicolumn{1}{l|}{Precision} & \multicolumn{1}{l|}{Recall} & \multicolumn{1}{l|}{$F_1$} & \multicolumn{1}{c|}{Precision} & \multicolumn{1}{l|}{Recall} & \multicolumn{1}{l}{$F_1$} \\ \hline
\multirow{2}{*}{XLM-R}                         & Single-task                                 & \multicolumn{1}{c|}{84.14}        & \multicolumn{1}{c|}{82.98}     & 83.55                      & \multicolumn{1}{c|}{79.00}        & \multicolumn{1}{c|}{77.96}     & 78.48                      \\ 
     & Monica et al \cite{dixit2020robust}       & \multicolumn{1}{c|}{84.84}        & \multicolumn{1}{c|}{83.36}     & 84.10                      &\multicolumn{1}{c|}{78.97}        & \multicolumn{1}{c|}{78.25}     & 78.58                      \\ 
& JointCapPunc  & \multicolumn{1}{c|}{\textbf{85.66}}        & \multicolumn{1}{c|}{\textbf{85.12}}     & \textbf{85.39}                      & \multicolumn{1}{c|}{\textbf{79.02}}        & \multicolumn{1}{c|}{\textbf{78.45}}     & \textbf{78.74}                      \\ \hline
\multirow{2}{*}{vELECTRA}                      & Single-task                                 & \multicolumn{1}{c|}{83.97} & \multicolumn{1}{c|}{83.10}& 83.53& \multicolumn{1}{c|}{79.35} & \multicolumn{1}{c|}{77.96}     & 78.65                     \\ 
     & Monica et al \cite{dixit2020robust}             & \multicolumn{1}{c|}{84.67}        & \multicolumn{1}{c|}{83.65}     &    84.16                & \multicolumn{1}{c|}{79.36}        & \multicolumn{1}{c|}{79.31}     & 79.33   \\
& JointCapPunc                                & \multicolumn{1}{c|}{\textbf{86.55}}        & \multicolumn{1}{c|}{\textbf{86.95}}     &    \textbf{86.75 }               & \multicolumn{1}{c|}{\textbf{80.79}}        & \multicolumn{1}{c|}{\textbf{79.43}}     & \textbf{80.10 }                     \\ \hline
\end{tabular}

\label{tab:result}
\end{table*}

\begin{table*}[]

\centering
\caption{Ablation study (in \%) on testset (the decrease in the F1 score of the modified models is indicated in bold)}
\vspace{-0.3cm}
\begin{tabular}{l|l|lll|lll}
\hline
\multirow{2}{*}{Component} & \multicolumn{1}{c|}{\multirow{2}{*}{Modification}} & \multicolumn{3}{c|}{Capitalization}                               & \multicolumn{3}{c}{Punctuation}                                  \\ \cline{3-8} 
                           & \multicolumn{1}{c|}{}                              & \multicolumn{1}{c|}{Precision} & \multicolumn{1}{c|}{Recall} & $F_1$ & \multicolumn{1}{c|}{Precision} & \multicolumn{1}{c|}{Recall} & $F_1$ \\  \hline
Full                       & None                                               & \multicolumn{1}{c|}{86.55}        & \multicolumn{1}{c|}{86.95}     & 86.75   &  \multicolumn{1}{c|}{80.79}          & \multicolumn{1}{c|}{79.43}       &  80.10  \\ 
Capitalization feature     & Removal                                            & \multicolumn{1}{c|}{85.40}          & \multicolumn{1}{c|}{80.99}       &  83.13 (\textbf{3.62}$\downarrow$) & \multicolumn{1}{c|}{80.52}          & \multicolumn{1}{c|}{77.80}       &  79.13 (\textbf{0.97}$\downarrow$)  \\ 
CRF layer                  & Softmax                                            & \multicolumn{1}{c|}{86.24}          & \multicolumn{1}{c|}{81.93}       &  83.98 (\textbf{2.77}$\downarrow$)  & \multicolumn{1}{c|}{79.42 }          & \multicolumn{1}{c|}{78.00}       &   78.71 (\textbf{1.39}$\downarrow$) \\
Contextual embedding       & Fasttext                                           & \multicolumn{1}{c|}{79.72}          & \multicolumn{1}{c|}{49.05}       &   60.73  (\textbf{26.02}$\downarrow$)& \multicolumn{1}{c|}{62.30}          & \multicolumn{1}{c|}{41.03}       &     49.47 (\textbf{30.63}$\downarrow$) \\ \hline
\end{tabular}
\label{tab:ablation}
\end{table*}

\subsection{Ablation Study}
We evaluate the contribution of individual components of our model compare with our best model that using vELECTRA as encoding layer and full components as follows:

\begin{itemize}

\item \textbf{Contextual embedding:} We replace contextual word embedding based on pre-trained transformers language model by context-independent word embedding. In our experiment, we use Fasttext \cite{bojanowski2017enriching}, a variant of Word2Vec \cite{Mikolov2013EfficientEO} which solved unknown words problem by using sub-word models.
\item \textbf{Capitalization feature:} We remove the capitalization feature and only use contextual embedding to predict punctuation mark.
\item \textbf{Conditional random fields:} We replace the CRF layer by softmax function for inferencing punctuation mark.
\end{itemize}
Table \ref{tab:ablation} shows the impact of all component that contributed to the success of our joint learning model.The performance of the system degrades when we modified or removed a component. While the full model got capitalization $F_1$ of 86.75\% and punctuation $F_1$ of 80.10\%, the capitalization $F_1$ and punctuation $F_1$ reduced to 83.13\%, 79.13\% respectively when we remove capitalization feature and only use contextual embedding to prediction punctuation mark. The $F_1$ score of capitalization prediction task and punctuation prediction task are only 83.98\%, 78.71\% respectively when we replaced the CRF layer by softmax function. Finally, we find that replacing contextual embeddings by independent word embedding (Fasttext) leads to the biggest decrease 23.43\% capitalization $F_1$ and 29.86\% punctuation $F_1$.  

\section{Conclusions}
In this paper, we presented the public dataset for Vietnamese capitalization and punctuation recovery. ViCapPunc dataset can be utilized to develop a text normalization module for ASR systems focus on medical domain which used to record the conversation between doctors and COVID-19 patient. In addition, we also proposed an effective model, namely JointCapPunc, for jointly learning capitalization and punctuation recovery. In particular, JointCapPunc based on transformers based pre-trained language models to learn essential words in the sentences. Soft capitalization embedding combined with word representation can produce useful features for punctuation prediction task. Therefor our model are able to capture the relationship between the two tasks. We conduct experiments and error analysis on our dataset and the results show that jointly learning model is more effective than baseline models (ie., single training approach, joint model Monica et al \cite{dixit2020robust}). However our model is based on transformers pre-trained language models that has a giant parameters scale, which might face difficulties in ASR pipeline due to the increase of latency.
For future work, we plan to optimize our model to smaller size for faster and lighter footprint.
\section{Acknowledgements}
We would like to thank AIC Technology Group, Hanoi, Vietnam for financial support which made this work possible.

\bibliographystyle{IEEEtran}

\bibliography{mybib}

\end{document}